\documentclass[12pt]{l4dc2020} 
\usepackage{graphicx}
\usepackage{wrapfig}
\usepackage{natbib}
\usepackage{graphicx}
\usepackage{amsmath}
\usepackage{float}
\usepackage{subcaption}
\usepackage[percent]{overpic}
\usepackage{multirow}
\usepackage{subcaption}
\usepackage[export]{adjustbox}
\allowdisplaybreaks
\usepackage{lmodern}
\usepackage{siunitx}
\usepackage{booktabs}
\usepackage{etoolbox}
\usepackage{hyperref}
\hypersetup{colorlinks,
            linkcolor=blue,
            citecolor=blue,
            urlcolor=magenta,
            linktocpage,
            plainpages=false}

\title[System Identification of Spring-Rod Systems with Differentiable Physics Engines]{A First Principles Approach for Data-Efficient System Identification\\ of Spring-Rod Systems via Differentiable Physics Engines}
\usepackage{times}

\author{%
 \Name{Kun Wang} \Email{kun.wang2012@rutgers.edu}\\
 \addr Robotics Lab, 1 Spring Street, New Brunswick, NJ 08901 
 \AND
 \Name{Mridul Aanjaneya} \Email{mridul.aanjaneya@rutgers.edu}\\
 \addr CBIM, 617 Bowser Rd, Piscataway, NJ 08854%
 \AND
 \Name{Kostas Bekris} \Email{kostas.bekris@cs.rutgers.edu}\\
 \addr Robotics Lab, 1 Spring Street, New Brunswick, NJ 08901
}
\vspace{-.3in}
\begin{document}
\vspace{-.3in}
\maketitle

\vspace{-.3in}

\begin{abstract}%
We propose a novel differentiable physics engine for system identification of complex spring-rod assemblies. Unlike black-box  data-driven methods for learning the evolution of a dynamical system \emph{and} its parameters, we modularize the design of our engine using a discrete form of the governing equations of motion, similar to a traditional physics engine. We further reduce the dimension from 3D to 1D for each module, which allows efficient learning of system parameters using linear regression. As a side benefit, the regression parameters correspond to physical quantities, such as spring stiffness or the mass of the rod, making the pipeline explainable. The approach significantly reduces the amount of training data required, and also avoids iterative identification of data sampling and model training. We compare the performance of the proposed engine with previous solutions, and demonstrate its efficacy on tensegrity systems, such as NASA's icosahedron.
\end{abstract}

\vspace{-.1in}
\section{Introduction}
\vspace{-.1in}


Performing experiments on real robots can be time-consuming, expensive, or dangerous. As such, it is often preferable to explore policies in simulation first, and then transfer them to the real robot. To minimize the reality gap, accurate system identification is important. The traditional approach \cite{swevers1997optimal, hansen2001completely} for system identification is to collect ground truth data and subsequently optimize the dynamics model over certain parameters of a physics simulation engine, so as to minimize the difference between the predicted trajectory and the ground truth. This process is repeated iteratively until the difference becomes small enough. Apart from being slow, this iterative process can potentially damage a vulnerable robot. As an alternative, recent approaches have been exploring the use of a \emph{differentiable} physics engine, such as a neural network, that would allow for parameter inference using backpropagation. Nevertheless, training a differentiable physics engine requires massive amounts of data. While there has been some work on methods for online system identification \cite{Yu-RSS-17,allevato2019tunenet}, which can learn parameters with much less real data, such methods are limited to a small number of parameters.


The problem of system identification becomes exacerbated for soft robots, which have infinite degrees of freedom. Physics-based methods for simulation require accurate models that capture non-linear material behavior, which are difficult to construct. In contrast, data-driven methods can simulate any system from observed data, with sufficient training data. But the large number of variables and non-linear material properties necessitate copious amounts of training data.

\textcolor{black}{Cable-driven robots are gaining increasing attention due to their adaptiveness and safety. Tensegrity structures have many applications: from manipulation \cite{lessard2016bio}, locomotion \cite{sabelhaus2018design}, morphing airfoil \cite{chen2020design} to spacecraft lander \cite{bruce2014superball}. While useful and versatile, they are difficult to accurately model and control. Identifying system parameters is necessary, either to learn controllers in simulation (as real-world experiments are time-consuming, expensive and dangerous), or for traditional model-based control. In all these cases, the spring-rod representation considered in this work is the basic modeling element. 
}

Motivated by these issues, we propose a data-driven differentiable physics engine that combines the benefits of data-driven and physics-based models, while alleviating most of their drawbacks, and is designed from first principles. Previous data-driven models have required large amounts of data, because they learn the  parameters \emph{and} the physics of the system.  Furthermore, the hidden variables and black box nature of these models are not explainable, and difficult to transfer to new environments. Our approach is based on the observation that the equations that govern the motion of such systems are well-understood, and can be directly baked into the data-driven model. Such a design can reduce demands on training data and can also generalize to new environments, as the governing principles remain the same. We further simplify the differentiable engine by using a modular design, which compartmentalizes the problem of learning the dynamics of the whole system to smaller well-contained problems. For each module, we also reduce the dimension from 3D to 1D, by taking advantage the properties of spring-rod systems, which allows for efficient parameter inference using linear regression. As a side benefit, the regression parameters correspond to physical quantities, such as the spring stiffness or the mass of the rod, making the framework explainable. A video accompanying this work is available  \href{https://rutgers.box.com/shared/static/i9vvxpc8152i5e0zg897fj47sanen7nj.mp4}{here}\footnote{{https://rutgers.box.com/shared/static/i9vvxpc8152i5e0zg897fj47sanen7nj.mp4}}.

\begin{figure}
\vspace{-0.7cm}
    \centering
    \includegraphics[height=0.25\linewidth, trim={6.5cm 5cm 2cm 1cm}, clip]{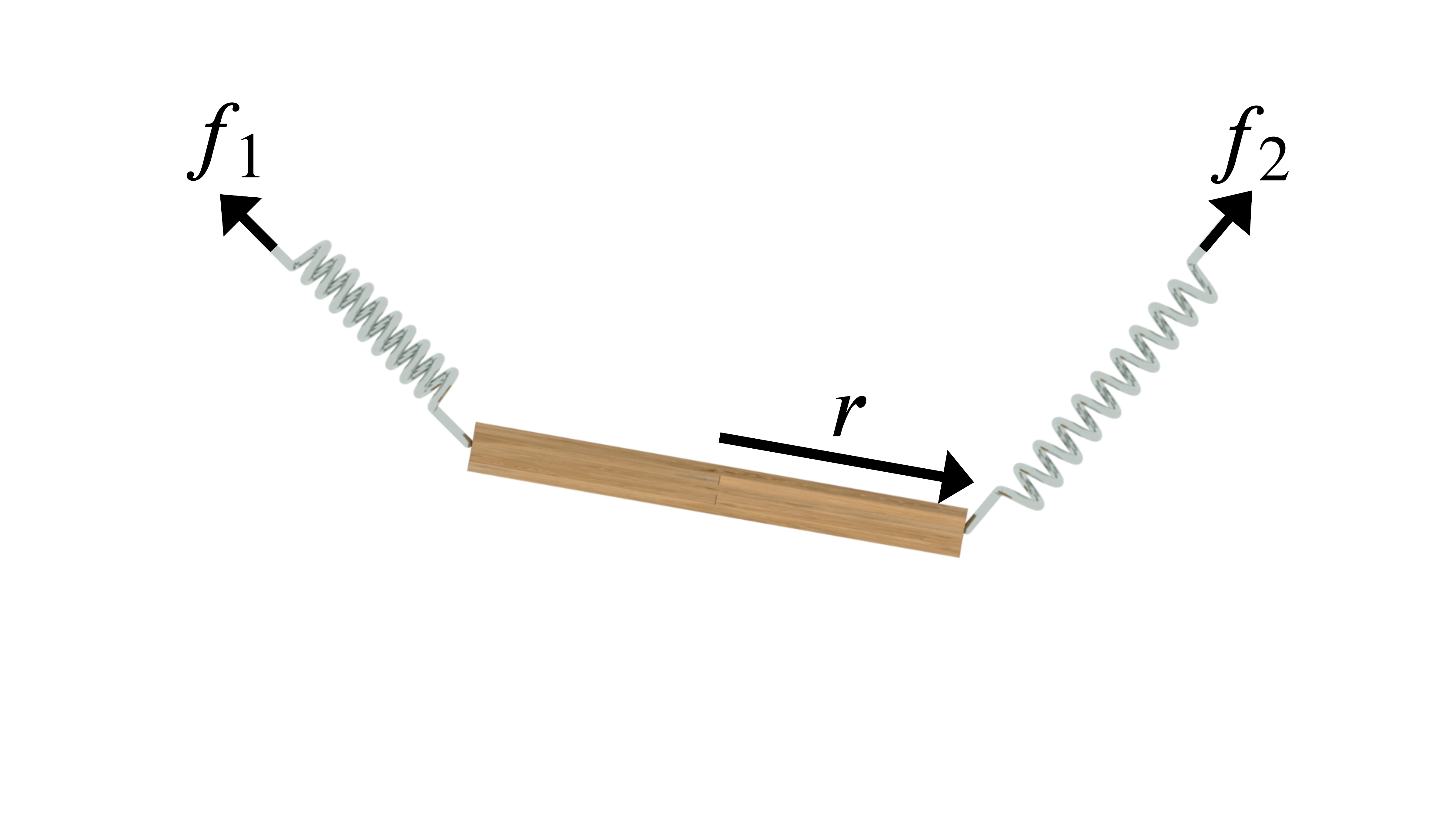}
    \includegraphics[height=0.25\linewidth, trim={0cm 0cm 0cm 0cm}, clip]{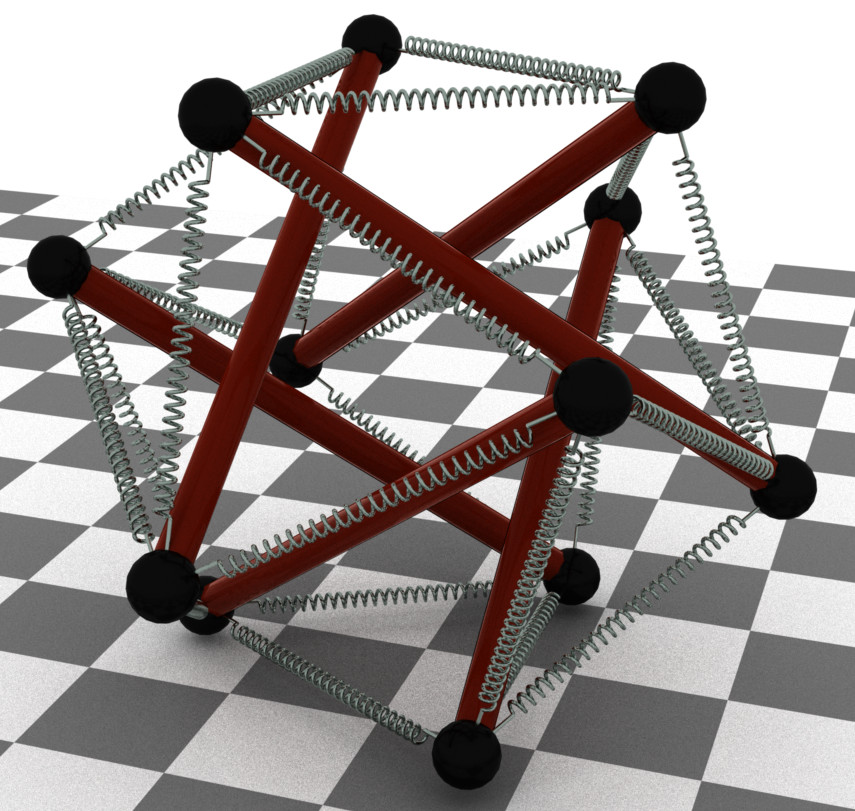}
    \includegraphics[height=0.25\linewidth]{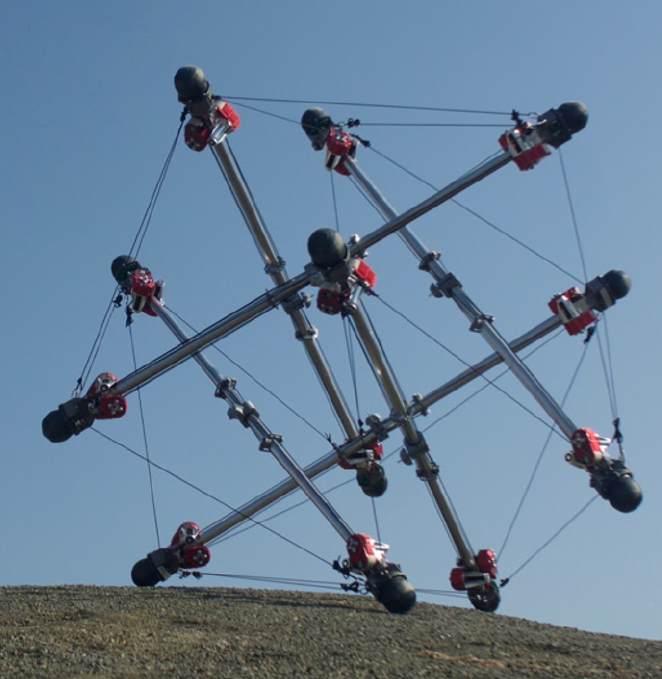}
    \vspace{-2mm}
    \caption{A basic element with one rod connected by two springs (left). A complex assembly of rods and springs forming a tensegrity robot in simulation (middle), and the real world (right).}
    \label{fig:spring_rod_model}
    \vspace{-5mm}
\end{figure}

\vspace{-3mm}
\section{Related Work}
\vspace{-.1in}
Traditional methods for system identification build a dynamics model by minimizing the prediction error~\cite{swevers1997optimal} \cite{hansen2001completely}. 
These methods require parameter refinement and data sampling in an iterative fashion, to decrease the prediction error. This iterative process can be avoided using data-driven techniques that directly fit a physics model to data~\cite{rosenblatt1958perceptron,rumelhart1986learning,asenov2019vid2param}. However, these techniques treat the dynamics as a black box, are data hungry, and require retraining in a new environment.


Instead of treating the environment as a black box, \cite{battaglia2016interaction} took the first step to modularize objects and their interactions in an \emph{interaction network}. Later, \cite{mrowca2018flexible} introduced a hierarchical relation network for graph-based object representation of rigid and soft bodies by decomposing them into particles.  Recently, \cite{li2019propagation} proposed the multi-step propagation network and \cite{NIPS2019_9672} applied a Hamiltonian network to conserve an energy-like quantity without damping. While these methods are an improvement over previous approaches, they still treat the interactions between different objects as black boxes and try to learn them from data, even though the governing equations of motion are well-understood.

Koopman operator theory from dynamical systems provides an alternative approach to learning the dynamics. The Koopman operator is a mechanism to ``lift'' lower dimensional non-linear features into higher dimensional linear features, which can subsequently be used to compute a linear dynamics model. \cite{bruder2019nonlinear} and \cite{li2019learning} have applied this technique to soft robot dynamics identification. However, the design of the Koopman operator is non-trival, and the high dimensional features are not explainable, making it challenging to estimate all physical parameters.

Quite a few authors have recently introduced differentiable physics engines that focus on many aspects not central to our work. For example, \cite{heiden2019interactive} predict forward dynamics of articulated rigid bodies, \cite{hu2019chainqueen} solve inverse problems using the Material Point Method (MPM), \cite{de2018end} address multi-body contact with linear complementarity problems (LCP), and \cite{landry2019differentiable} treat nonlinear optimization with the augmented Lagrangian method. To provide a general interface, \cite{hu2019difftaichi} introduced a compiler for differentiable programming. \cite{sain} use differentiable engines with traditional physics simulators for control. Researchers have also proposed differentiable engines specific to certain kinds of objects, such as molecules~\cite{jaxmd2019}, fluids~\cite{spnets2018}, and cloth~\cite{liang2019differentiable}. Our work on spring-rod systems is motivated by our recent work on tensegrity robots~\cite{surovik2019adaptive, littlefield2019kinodynamic, surovid2018any}.

\vspace{-3mm}
\section{Methods}
\vspace{-0.1in}

\begin{figure} 
\vspace{-0.7cm}
    \centering
    \includegraphics[width=\textwidth]{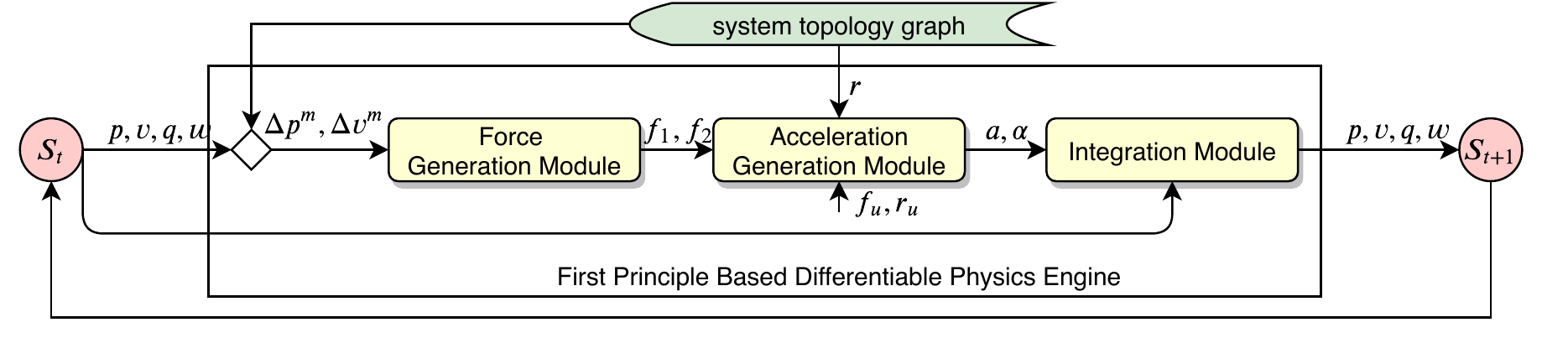}
    \vspace{-8mm}
    \caption{Flow chart showing the data flow when simulating one time step with our physics engine.}
    \label{fig:architecture}
    \vspace{-5mm}
\end{figure}

Our system views a spring-rod system as a composition of basic \emph{elements} (see Fig.~\ref{fig:spring_rod_model}(left)), where springs generate forces that influence rod dynamics. We subdivide each time step of the simulation into three modules: force generation, acceleration computation, and state update/integration (see Fig.~\ref{fig:architecture}). The physics engine takes as input the current rod state $S_t = \{p, v, q, \omega\}$, where $p$ is position, $v$ is linear velocity, $q$ is orientation (expressed as a quaternion), and $\omega$ is the angular velocity. Based on $S_t$, the position and linear velocity $p^{m}, v^{m}$ of the two rod endpoints is computed, and is used to compute the relative compression (or expansion) and velocity $\Delta p^{m}, \Delta v^{m}$ of the two attached springs. Then, the first module predicts the spring forces $f$, the second module computes the linear and angular accelerations $a,\alpha$, and the third module computes the new state $S_{t+1}$.


\vspace{-2mm}
\subsection{System Topology Graph}
\vspace{-2mm}

\begin{wrapfigure}[4]{r}{0.38\textwidth} 
\vspace*{-4mm}
\begin{overpic}[width=0.38\textwidth]{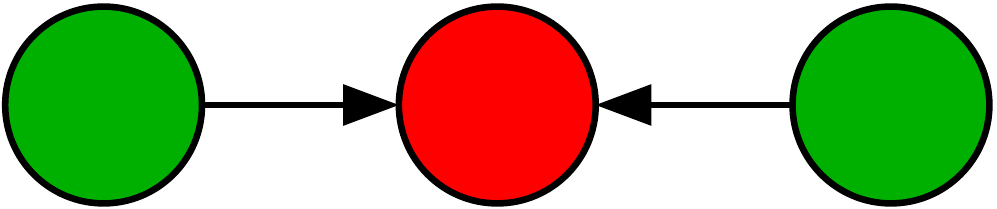}
    \put(1.5,8){{\small spring1}}
    \put(26,13){{$f_1$}}
    \put(45,8){{\small Rod}}
    \put(69.5,13){{$f_2$}}
    \put(80.5,8){{\small spring2}}
\end{overpic}
\vspace*{-8mm}
\caption{\small Element topology graph.}
\label{fig:topology}
\end{wrapfigure}
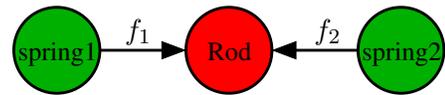
We use a topology graph to represent interconnections between different components of the spring-rod system. Each rod and spring has a corresponding vertex, and directed edges represent relations between them. Figure~\ref{fig:topology} shows an example topology graph for the basic spring-rod element shown in Figure~\ref{fig:spring_rod_model}(left).
Unlike~\cite{battaglia2016interaction}, who assumed all forces are applied at the center of mass of a rod, we apply forces at the endpoints $p^{m}$ of each rod, which is critical for accurate angular momentum computation.



\vspace{-2mm}
\subsection{Force Generation Module}
\vspace{-2mm}

The relative compression (or expansion) $\Delta p^{m}$ and velocity $\Delta v^{m}$ of each spring is given as input to the force generation module, which computes the spring forces $f$ as per the equations below:

\begin{figure}
    \centering
    \includegraphics[width=0.8\textwidth]{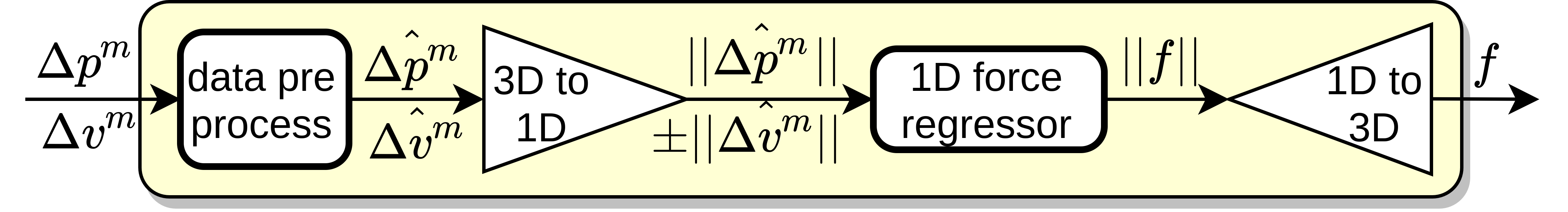}
    \vspace{-4mm}
    \caption{Force generation module, which uses dimensionality reduction to compute spring forces.}
    \label{fig:spring_physics_engine}
 \vspace{-2mm}
\end{figure}

 \vspace{-0.5cm}
\begin{align}
\hat{\Delta p^{m}} &= \Delta p^{m} - \Delta p^\text{rest}, \quad& \hat{\Delta v^{m}} &= (\Delta v^{m} \cdot \Delta p^{m}) \cdot (\Delta p^{m} / ||\Delta p^{m}||) \label{eq:pre-process}\\
    ||f|| &= - K ||\hat{\Delta p^{m}}|| \mp c ||\hat{\Delta v^{m}}||, \label{eq:hook} \quad& f &= ||f|| \cdot (\Delta p^{m} / ||\Delta p^{m}||)
\end{align}
where $K$ is the spring stiffness, $c$ is the damping parameter, and $\Delta p^\text{rest}$ is the spring rest length. To exploit the one-dimensional nature of each spring, we reduce $\Delta p^{m}$ and $\Delta v^{m}$ from 3D to 1D along the spring direction. This has the benefit of leaving only two unknown parameters, $K$ and $c$, which can be easily learned using linear regression in a data-efficient fashion. After computing the 1D spring force $f$, we project it back to 3D by multiplying with the normalized spring direction vector.


\vspace{-2mm}
\subsection{Acceleration Generation Module}
\vspace{-2mm}

\begin{figure}
    \centering
    \includegraphics[width=0.9\textwidth]{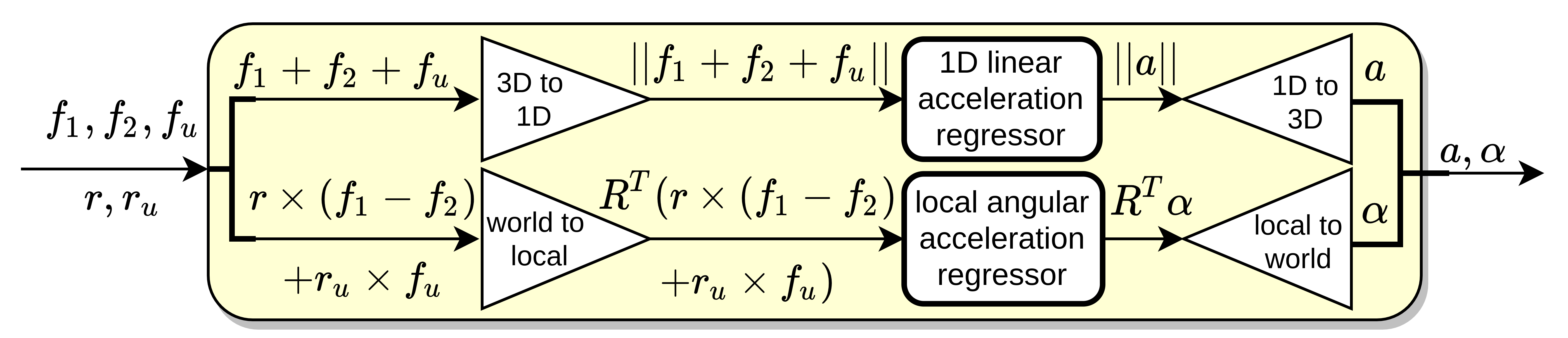}
    \vspace{-4mm}
    \caption{Acceleration generation module, which uses dimensionality reduction similar to the force generation module to compute rod accelerations, given spring forces at the two endpoints as input.}
    \label{fig:rod_physics_engine}
    \vspace{-5mm}
\end{figure}
The spring forces $f$ and control force $f_u$ are given as input to the acceleration generation module, which computes the linear and angular accelerations $a,\alpha$ of each rod as per the equations below:

\vspace{-0.5cm}
\begin{align}
    &||f_1 + f_2 + f_u|| = M ||a||,&  & R^{T}(r \times (f_1 - f_2) + r_u \times f_u) = I(R^{T} \alpha) \label{eq:2} \\
    &a = ||a|| \cdot \dfrac{f_1 + f_2 + f_u}{||f_1 + f_2 + f_u||},&  & \alpha =RI^{-1}R^{T} (r \times (f_1 - f_2) + r_u \times f_u) \label{eq:3}
    \vspace{-3mm}
\end{align}
\textcolor{black}{where $f_1$ and $f_2$ are spring forces on the two rod ends, $f_u$ is control force, $r$ is the half-length rod vector, $r_u$ is control force arm, $R$ is the rod local/world frame rotation matrix, $M$ is the rod mass and $I$ is the local frame moment of inertia of the rod. $M$ and $I$ are unknown parameters to identify. In a complex system where a rod is connected with multiple springs on one (or both) of its endpoints, $f_1$ and $f_2$ denote the \emph{aggregate} spring force. Analogous to the spring case, this work exploits the 1D nature of each rod by computing the norm of the cumulative force $f_1 + f_2 + f_u$, reducing the dimension from 3D to 1D. After computing the 1D linear acceleration $a$, the next step is to map them back to 3D using equation (\ref{eq:3}). We compute angular acceleration $R^{T}\alpha$ in local frame, where $I$ is a diagonal 3x3 matrix because of the symmetry of a rod. This allows to represent it as a 3x1 vector instead, requiring less data for regression.}


\vspace{-2mm}
\subsection{Integration Module and Method Implementation}
\label{integraion_module}
\vspace{-2mm}

The integration module computes forward dynamics of each rod using the current accelerations $a,\alpha$. We apply the semi-implicit Euler method~\cite{stewart2000implicit}
to compute the updated state $S_{t+1}=\{p_{t+1}, v_{t+1}, q_{t+1}, \omega_{t+1}\}$ at the end of the current time step.

\textcolor{black}{The learning module receives the current state $S_t$ and returns a prediction $\hat{S_{t+1}}$.  The loss function is the MSE between the predicted $\hat{S_{t+1}}$ and ground truth state $S_{t+1}$. The proposed decomposition, first-principles approach and the cable’s linear nature allow the application of linear regression, which helps with data efficiency. This linear regression step has been implemented as a single layer neural network without activation function on pyTorch \cite{paszke2019pytorch}. We trained 30 epochs with an Adam optimizer by a learning rate starting from 0.1 and reducing 50\% every 3 epochs.}

\vspace{-3mm}
\section{Experiments}
\vspace{-.1in}

We use two setups: 1) a simple spring-rod system (Fig. \ref{fig:spring_rod_model}(a)) and 2) a complex tensegrity system (Fig. \ref{fig:spring_rod_model}(b)).  For the tensegrity, we first assume uniform parameters for all rods and springs and then non-uniform ones, which is more realistic. The comparison methods represent two categories: 1) techniques that treat the system as a black box, such as Least Square (LS) optimization (L-BFGS-B: \cite{swevers1997optimal}), Covariance Matrix Adaptation Evolution Strategy (CMA-ES: \cite{hansen2001completely}), 2) methods that reason about physics and system topology, such as the Interaction Network \cite{battaglia2016interaction} and Koopman operator theory \cite{koopman1931hamiltonian}, which have been used for dynamics model identification \cite{bruder2019nonlinear} \cite{li2019learning} \cite{abraham2017model}. 
\textcolor{black}{All these approaches get access to the system state along executed trajectories, i.e., the 3D position, 3D linear velocity, quaternion and 3D angular velocity of each rigid element. This work does \emph{not} use any additional information, such as spring forces, accelerations, etc. The task is to estimate system parameters including spring stiffness $K$, damping $c$ and rod mass $M$, inertia $I$.}

\textcolor{black}{These comparison algorithms were optimized to improve performance on the target experiments. For LS and CMA-ES: the search was constrained to a reasonable range, which increases success ratio. The CMA tolerance was adjusted to 1, which allows to tune the trade-off between accuracy and speed. For the Interaction Network: instead of using the raw object state as in the original paper, the performed experiments used the relative position and velocity corresponding to spring-based models, which is easier for the neural network to learn. For the Koopman operator: instead of  basis functions like sinusoids/exponentials, a physics and topology-aware operator was used. All these adaptations increased performance. The methods were tested on an Intel Core i7 6700K CPU, 32G DRAM, 1T SSD and a GeForce GTX TITAN X Pascal GPU.}

\vspace{-.1in}
\subsection{Simple Spring-Rod System Identification}
\vspace{-.1in}
This system contains 2 springs and 1 rod as in Fig. \ref{fig:spring_rod_model}(a). Each spring is attached to one end of a rod and a fixed nail. The system is modeled in MuJoCo \cite{todorov2012mujoco}. Each rod is 2 meters long. The rest length of each spring is 1 m and 1.414 m. Gravity is ignored in this case.\\
\vspace{-.1in}


\noindent {\sc Black Box System Identification} \textcolor{black}{Comparisons to black box identification methods are included since one of this paper's objectives is to emphasize the limitations of such increasingly popular black-box models. Note that the black-box models have full access to the dynamics, as they resample a new trajectory from the simulator after each optimization iteration. The proposed method doesn't. It instead integrates knowledge from first principles into the optimization process of a differentiable physics engine.} We sample 100 trajectories with different initial conditions as ground truth data in order to train the methods. Each trajectory is 2000 time steps long.



\begin{figure}[htb]
\vspace{-12pt}
\begin{minipage}[c]{.4\linewidth}
\vspace{0pt}
\centering
\includegraphics[width=\linewidth]{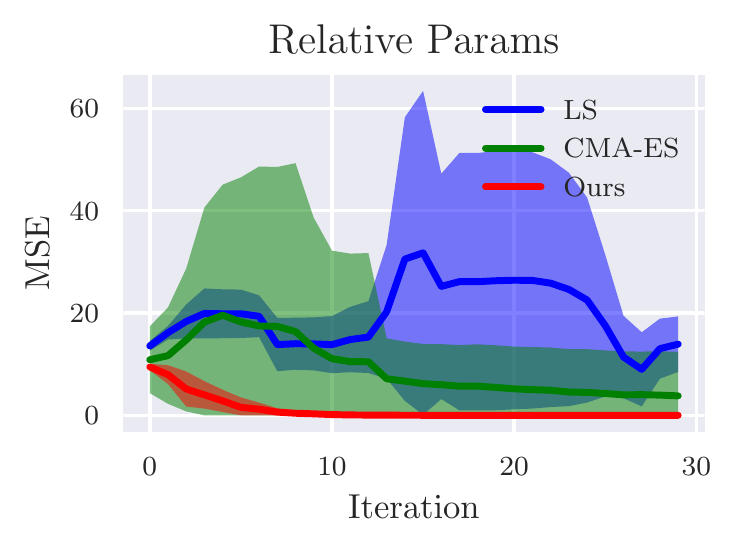}
\end{minipage}%
\begin{minipage}[c]{.59\linewidth}
\vspace{0pt}
\centering
\small

\sisetup{detect-weight,mode=text}
\renewrobustcmd{\bfseries}{\fontseries{b}\selectfont}
\renewrobustcmd{\boldmath}{}
\newrobustcmd{\B}{\bfseries}
\addtolength{\tabcolsep}{-4.1pt}

\begin{tabular}{ c c c c c}
method & $K/M$ & $c/M$ & sec/itr & success\% \\
 \hline
GT & 10 & 1 \\ 
LS & 19.19$\pm$7.91 & 1.51$\pm$1.16 & 1.97$\pm$1.43 & 0.01 \\
CMA-ES & 10.29$\pm$3.73 & 3.61$\pm$8.22 & 2.25$\pm$0.30 & 0.18 \\
Ours & \textbf{10.00$\pm$0.01} & \textbf{1.00$\pm$0.00} & \textbf{0.11$\pm$0.01} & \textbf{1.00} \\
\end{tabular}
\end{minipage}
\vspace{-0.5cm}
\caption{(left) mean square error after $N$ iterations for the simple rod-spring systeml; (right) parameters estimated after 30 iterations. The proposed approach is the most accurate and the fastest.}
\label{fig:harder_simple_system_result}
\vspace{-0.15in}
\end{figure}

\textcolor{black}{The task is to estimate stiffness $K$, damping $c$ and mass $M$, which are set to 100, 10 and 10 for ground truth. To assist the optimizer of LS and CMA-ES, we assume inertia $I$ is inferred from the rod geometry and constrain the range of solution to a positive interval $[0.1, 1000]$ for $K>0, c>0$, whilst the proposed approach doesn't benefit. From equations \eqref{eq:hook} \eqref{eq:2}, we can infer that:
$a = -(K/M)(\Delta x_1 + \Delta x_2) - (c/M) (\Delta v_1 + \Delta v_2).$
Thus, any parameter combination satisfying $K/M = 100/10=10$ and $c/M = 10/10 = 1$ is a solution. Instead of evaluating the absolute parameters $K, c$, we test the relative ratios $K/M$ and $c/M$.  MSE in Fig.\ref{fig:harder_simple_system_result} means difference in relative parameters $K/M, c/M$. Success is achieved when the parameter error within 5\% of the ground truth. The proposed approach works best whilst both LS and CMA-ES failed in most of the cases as this problem is non-convex and has infinite number of possible answers.} The additional blackbox experiments can be found in appendix \ref{apx:black_box_identification}.

\begin{figure}[t]
\vspace{-0.3cm}
    \centering
    \begin{minipage}[b]{.32\linewidth}
    \includegraphics[height=1.4cm, trim={1.8cm 0cm 0cm 0cm}, clip, left]{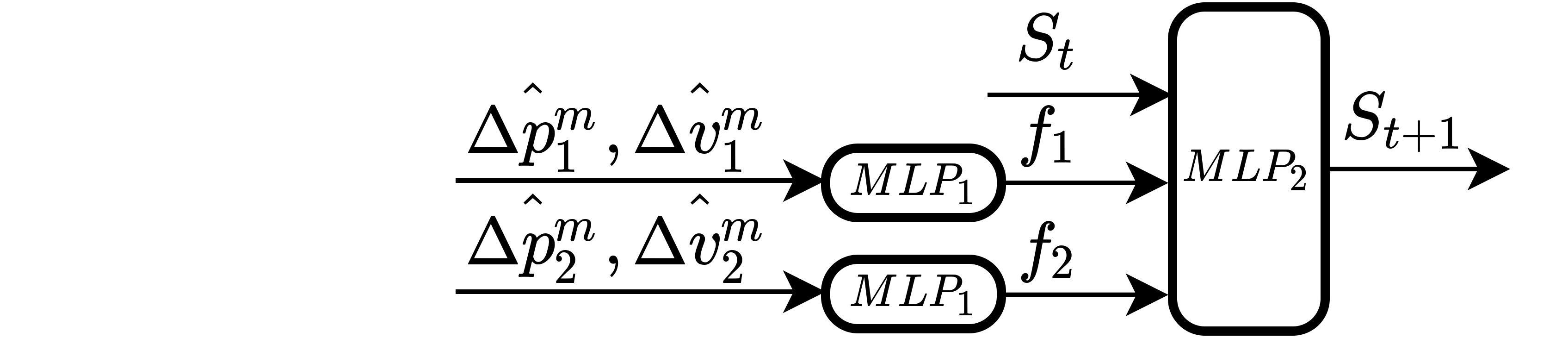}
    \vspace{-7mm}
    \caption{Interaction Network}
    \label{fig:interaction_network_architecture}
    \end{minipage}%
    \hfill 
    \begin{minipage}[b]{.67\linewidth}
    \includegraphics[width=\textwidth]{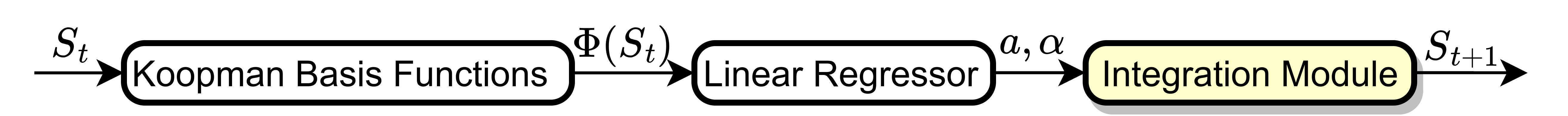}
    \vspace{-7mm}
    \caption{Koopman with Integration Module}
    \label{fig:koopman_w_int_architecture}
    \label{fig:interaction_network_w_int_architecture}
    \end{minipage}
\vspace{-0.2in}
\end{figure}

\noindent {\sc Physics and System Topology Aware Identification} We consider alternatives, which do reason about physical properties. \textbf{Interaction} is an improved version of the Interaction Network \cite{battaglia2016interaction} as shown in Fig. \ref{fig:interaction_network_architecture}. It has two Multilayer Perceptrons (MLPs), one to generate spring forces $f$ and the other to generate rod state $S_{t+1}$. Unlike \cite{battaglia2016interaction}, which takes raw state $S_t$ as input, we apply equation \eqref{eq:pre-process} to generate $\hat{\Delta p^{m}_t}, \hat{\Delta v^{m}_t}$ as input. \textbf{Interaction+Int} appends the integration module to the Interaction Network, and replaces input $S_t$ by $r$.


\begin{wrapfigure}{l}{0.5\textwidth}
    \begin{minipage}[t]{\linewidth}
    \vspace{0pt}
    \centering
    \includegraphics[width=0.49\linewidth]{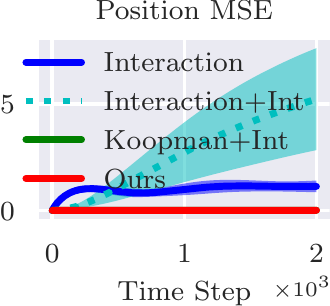}
    \includegraphics[width=0.49\linewidth]{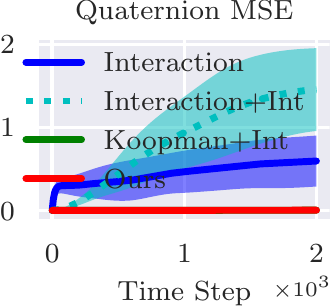}
    \vspace{-0.8cm}
    \caption{Physics and Topology-aware Identification}
    \label{fig:simple_system_known_topology_mse}
    \end{minipage}
    \vspace{-0.5cm}
\end{wrapfigure}

Recently, researchers focus on the Koopman operator to fit dynamical models. It applies the kernel trick, similar to SVMs, to map nonlinear features to many dimensions, where we can find linear relations. Instead of kernels, we construct Koopman operators based on the physical properties. We apply the following  polynomial basis functions as vector-valued functions to generate the approximate Koopman operator: $\Phi(X) = [1, r, \hat{\Delta p_{1}^m}, \hat{\Delta p_{2}^m}, \hat{\Delta v_{1}^m}, \hat{\Delta v_{2}^m}, \phi_i], \quad$ where $\phi_i = a^{\alpha_i} b^{\beta_i},$ $a, b \in [r_j, (\hat{\Delta p_{k}^m})_j, (\hat{\Delta v_{k}^m})_j ]$ and $j \in \{x, y, z\}$, $k \in \{1, 2\}$.The terms $\alpha_i$ and $\beta_i$ are non-negative integers, index $i$ tabulates all the combinations such that $\alpha_i +\beta_i \leq Q$ and $Q > 1$ defines the largest allowed polynomial degree. We define $Q = 2$. Detailed derivation can be found in Appendix \ref{apx:koopman_derivation}. We only use the Koopman operator to predict accelerations and apply the Integration Module to map them to $S_{t+1}$, which helps to reduce the complexity of the basis functions. We denote the approach as \textbf{Koopman+Int} (Fig. \ref{fig:koopman_w_int_architecture}). \textcolor{black}{We sampled 1000 trajectories, lasting 2,000 time steps, with different initial conditions for training and 200 for validation, and 100 for testing. Trajectories are converted to $(S_t, S_{t+1})$ pairs for training. Position/Quaternion MSE in Fig.\ref{fig:simple_system_known_topology_mse} means accumulated trajectory difference.} \textbf{Interaction} only predicts a $S_{t+1}$ in training data that is close to $S_t$. \textbf{Interaction+Int} experiences increasing error from accumulated prediction errors. The Koopman operator \textbf{Koopman+Int} designed from first principles gives accurate predictions similar to \textbf{Ours} in this simple system.



\vspace{-.1in}
\subsection{Complex Tensegrity Model Identification} 
\vspace{-.1in}
We consider an icosahedron tensegrity system as shown in Fig. \ref{fig:spring_rod_model} (c). It is composed of 6 rods and 24 springs. Each rod is connected to 8 springs and has a length of 1.04m. Each spring's rest length is 0.637m. We set the gravity constant to $g=-9.81$ in Mujoco. We collect 1000 trajectories with different initial conditions for training, 200 for validation and 100 for testing. Since only Koopman operator performs well in the simple system, it is the comparison point in this complex setup.



\begin{figure}[h]
    \begin{minipage}[t]{.5\linewidth}
    \vspace{0pt}
    \centering
    \includegraphics[width=0.49\linewidth]{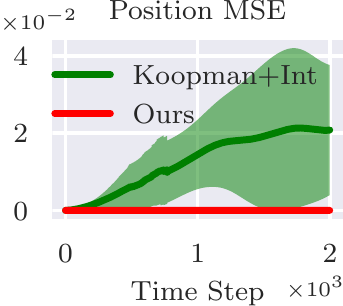}
    \includegraphics[width=0.49\linewidth]{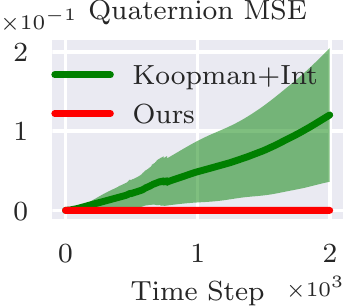}
    \vspace{-0.7cm}
    \caption*{a) Uniform Parameters}
    \end{minipage}%
    \begin{minipage}[t]{.5\linewidth}
    \vspace{0pt}
    \centering
    \includegraphics[width=0.49\linewidth]{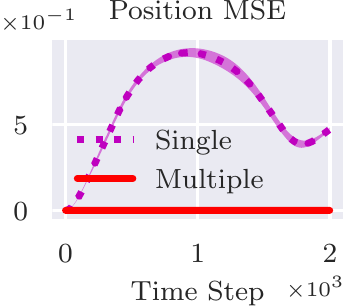}
    \includegraphics[width=0.49\linewidth]{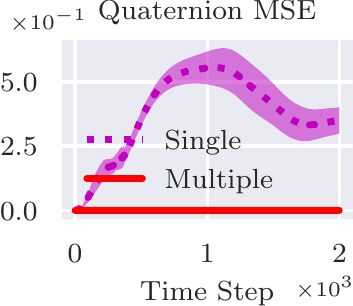}
    \vspace{-0.7cm}
    \caption*{b) Different Parameters}
    \end{minipage}
    \vspace{-0.4cm}
    \caption{Comparison with the Koopman approach on the complex tensegrity system.}
    \label{fig:superball_system_identification}
    \vspace{-0.5cm}
\end{figure}

\noindent {\sc Tensegrity with Uniform Parameters} \label{sec:tensegrity_same_params} Initially, we set uniform parameters for all rods and springs, i.e. set the mass of rod to 10, and set stiffness and damping to 100 and 10. Thus we keep the number of parameters to estimate, but increase the complexity of the environment. The structure of the Koopman operator is the same as before. The result is shown in Figure \ref{fig:superball_system_identification}. \textbf{Our} approach outperforms \textbf{Koopman+Int} because designing basis functions for The Koopman operator has an increased data requirement relative to our approach.



\noindent {\sc Tensegrity with Non-Uniform Parameters} On a real platform the parameters of each element are not the same. We added Gaussian noise to each parameter with $\mu=0, \sigma=0.2*\textit{parameter}$ resulting in 54 parameters for rod masses, spring stiffness and damping. To deal with this, we use individual regressors for each of rod and spring, which can adapt parameter divergence and achieve higher accuracy. Fig. \ref{fig:superball_system_identification} b) compares two versions of our approach. The \textbf{Multiple} uses individual regressors to achieve lower error compared relative to the \textbf{Single} regressor. The errors reduce at later time steps because of the periodical property of spring oscillation. 

\vspace{-.1in}
\subsection{Data Efficiency Experiment}
\begin{figure}[h]
\vspace{-.1in}
\begin{minipage}[t]{.5\linewidth}
\vspace{0pt}
\centering
\includegraphics[width=0.49\linewidth]{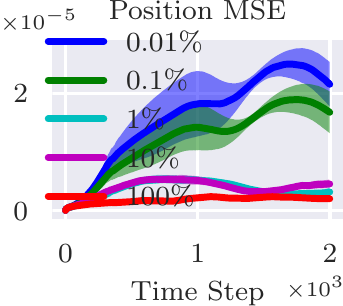}
\includegraphics[width=0.49\linewidth]{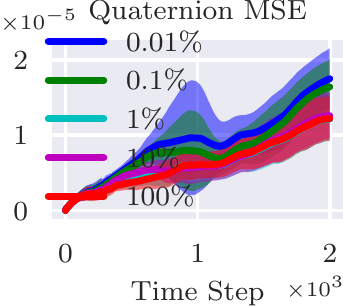}
\vspace{-0.7cm}
\caption*{a) Efficiency on Various Training Set}
\end{minipage}%
\begin{minipage}[t]{.5\linewidth}
\vspace{0pt}
\centering
\includegraphics[width=0.49\linewidth]{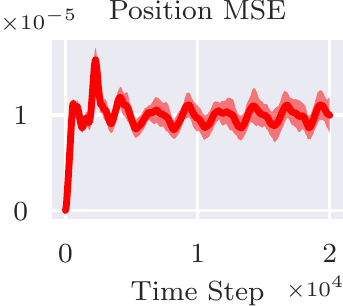}
\includegraphics[width=0.49\linewidth]{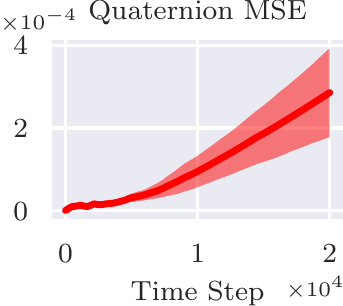}
\vspace{-0.7cm}
\caption*{b) Generalization on New Testing Set}
\end{minipage}
\vspace{-0.4cm}
\caption{Data Efficiency and Model Generalization Experiment.}
\label{fig:superball_diff_params_system_data_efficient_and_generalization}
\vspace{-0.2in}
\end{figure}

\vspace{-.1in}
The proposed method has relatively small data requirements as shown in Fig. \ref{fig:superball_diff_params_system_data_efficient_and_generalization} a). Instead of training on 1000 trajectories, which have 736,167 time steps in total, we train our model with less data and evaluate performance. We randomly select 10\%, 1\%, 0.1\%, 0.01\%  of the 736,167 time steps for training. The model achieves good performance even with 73 time steps for training. All trajectories are from the complex tensegrity setup with different parameters. We performed similar experiments on the Koopman operator, which can be found in Appendix \ref{apx:koopman_efficiency}. 


\textcolor{black}{The proposed solution achieves very low error at a magnitude of $10^{-5}$, since it: 1) introduces a first-principles approach in learning physical parameters (compared against the popular Interaction Network); 2) removes redundant data from regression (compared to the Koopman operator); 3) operates -for now- over relatively clean data from simulation before moving to real-world data. }

\vspace{-.1in}
\subsection{Model Generalization Experiment}
\vspace{-.1in}
\begin{wrapfigure}{r}{0.5\textwidth}
    \vspace{-.3in}
    \centering
    \includegraphics[width=0.49\textwidth]{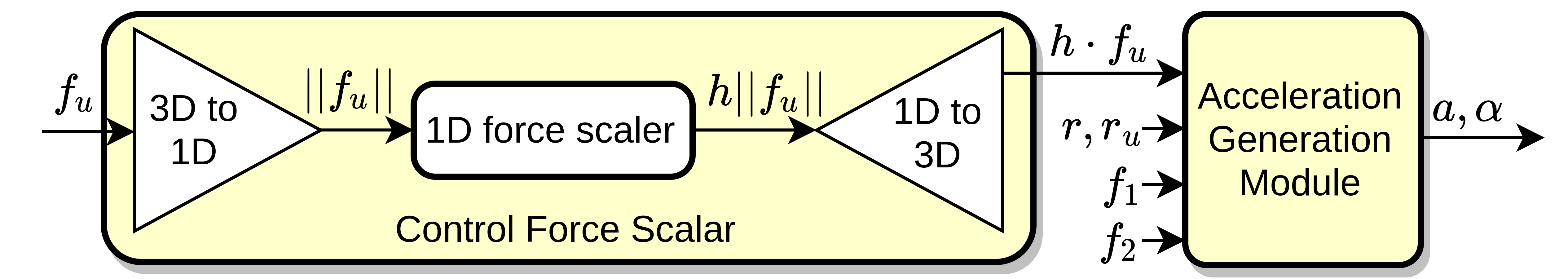}
    \vspace{-.15in}
    \caption{Control Force Scalar}
   \label{fig:control_force_scalar}
 \vspace{-0.2in}
\end{wrapfigure}
\textcolor{black}{This section generalizes the physics engine trained with a dataset without external forces to a dataset with such forces.} We are interested in evaluating: 1) how the physics engine performs for longer time horizons (e.g., after 2000 time steps); 2) if it can adapt to new scenarios. We generate a new test set for the complex tensegrity setup with different parameters: 100 trajectories lasting 20,000 time steps. We also add a random directed perturbation force $f_u$ to a random selected rod every 100 time steps. \textcolor{black}{The external force $f_u$ does not have the same scale as the internal spring forces, so we add a new module that aims to account for the external force}, i.e. the control force scalar module, containing only one parameter $h$, as in Fig. \ref{fig:control_force_scalar}.  We also apply dimensionality reduction to improve data efficiency. \textcolor{black}{The tuning process is to freeze all other modules' weights and train with object states from the new dataset. The module converges to a stable value.} The error graphs are shown in Fig. \ref{fig:superball_diff_params_system_data_efficient_and_generalization} b). The results on a third data set (with 4,000 time steps) is available in Appendix \ref{apx:generalization_4000}. 

\vspace{-.1in}
\section{Conclusion and Future Work}
\vspace{-.1in}
This paper proposes a differentiable physics engine for system identification of spring-rod systems based on first principles. The engine has three modules: force generation, acceleration generation and integration, which express the corresponding physical processes of spring-rod systems. This results in reduced data requirements and improved parameter accuracy. It also provides an explainable, accurate and fast physics engine. In the future, we plan to address contacts and friction. This will involve replacing the linear regressor with nonlinear models in the existing modules. To overcome noise in real data, we plan the addition of a residual network along with the nonlinear model. These changes may also help with temporal scalability.


\noindent {\bf Acknowledgment:} Chengguizi Han for rendering the different systems in the figures and video, and Craig Schroeder for sharing his ideas about photorealistic rendering  of springs. This work was supported by NASA ECF grant NNX15AU47G, NSF award 1723869, Rutgers University start-up grant, and the Ralph E. Powe Junior Faculty Enhancement Award. Any opinions and conclusions expressed in this work are made by the authors and do not necessarily reflect the views of the sponsor.

\bibliography{references}

\pagebreak

\appendix
\section{Simple Spring-Rod System Black Box Identification} \label{apx:black_box_identification}
\begin{figure}[htb]
\vspace{-12pt}
\begin{minipage}[c]{.4\linewidth}
\vspace{0pt}
\small
\includegraphics[width=\linewidth]{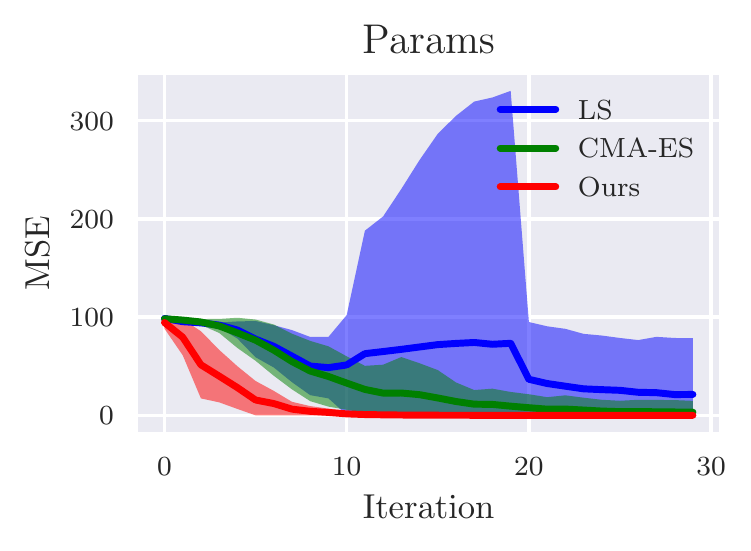}
\end{minipage}%
\hfill
\begin{minipage}[c]{.59\linewidth}
\vspace{0pt}
\centering
\small

\sisetup{detect-weight,mode=text}
\renewrobustcmd{\bfseries}{\fontseries{b}\selectfont}
\renewrobustcmd{\boldmath}{}
\newrobustcmd{\B}{\bfseries}
\addtolength{\tabcolsep}{-4.1pt}

\begin{tabular}{c c c c c}
method & stiffness $K$ & damping $c$ & sec/iter & success\%\\
 \hline
GT & 100 & 10 \\ 
LS & 98.02$\pm$50.71 & 51.49$\pm$247.58 & 1.32$\pm$3.14 & 0.42 \\
CMA-ES & 102.35$\pm$16.73 & 11.01$\pm$5.88 & 1.94$\pm$0.22 & 0.92 \\
Ours & \textbf{100.00$\pm$0.07} & \textbf{10.00$\pm$0.01} & \textbf{0.11$\pm$0.01} & \textbf{1.00} \\
\end{tabular}
\end{minipage}
\vspace{-0.5cm}
\caption{(left) mean square error after $N$ iterations for the simple rod-spring system; (right) parameters estimated after 30 iterations. The proposed approach is the most accurate and the fastest.}
\label{simple_system_blackbox_result}
\vspace{-0.15in}
\end{figure}

\paragraph{Comparison with Numerical Optimization Methods} We assume the rod mass is known, i.e. $M=10kg$, and only try to identify spring stiffness $K$ and damping $c$. The ground truth stiffness and damping are set to $K=100$ and $c=10$ respectively. \textcolor{black}{For each method, a ground truth trajectory is used as reference and then the mean and standard deviation are computed for robustness evaluation.} To assist the optimizer of LS and CMA-ES, we constrain the range of solution to a positive interval $[0.1, 1000]$ for $K>0, c>0$. Success is achieved when the parameter error within 5\% of the ground truth. \textcolor{black}{MSE in Fig.\ref{simple_system_blackbox_result} means difference in parameters $K, c$.}

The initial guess for LS and CMA-ES is 1. The initial parameters in our physics engine are also set to $1$. The CMA tolerance is set to 1. Our approach can converge to a precise model within 10 optimization iterations and takes 1.1s. LS has a large variance as it often doesn't converge. CMA-ES is more stable but still worse than our solution.

\begin{figure}[h]
\vspace{-0.5cm}
\centering
    \begin{minipage}[c]{0.5\linewidth}
        \vspace{0pt}
        \centering
        \includegraphics[width=0.5\linewidth]{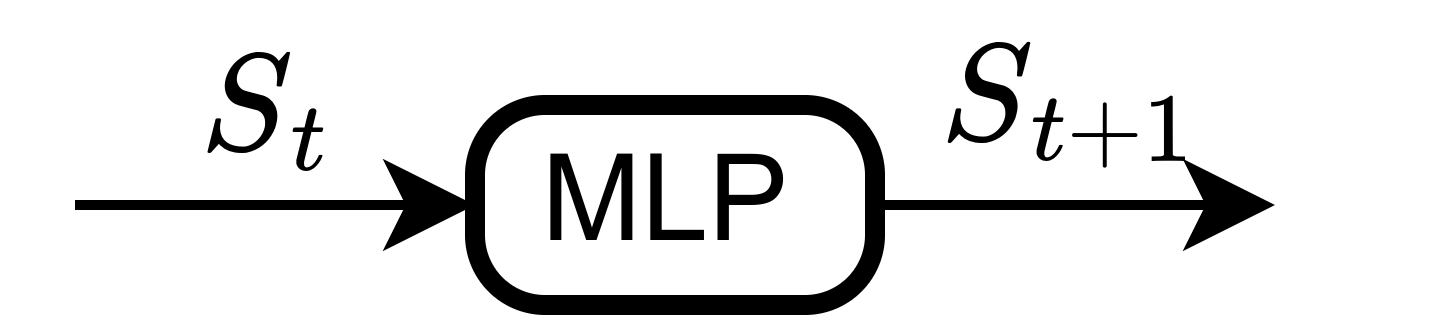}
        \includegraphics[width=\linewidth]{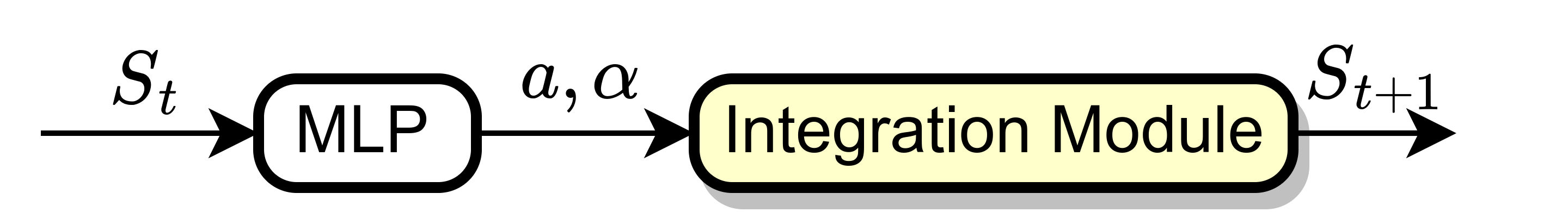}
    \end{minipage}%
    \begin{minipage}[c]{0.5\linewidth}
    \vspace{0pt}
    \centering
    \includegraphics[width=0.49\linewidth]{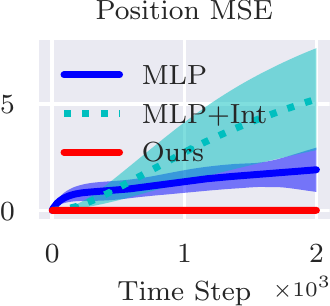}
    \includegraphics[width=0.49\linewidth]{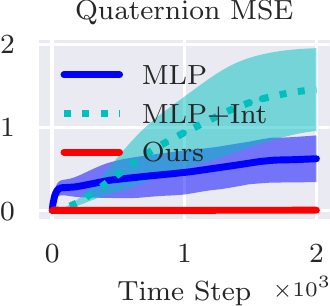}
    \end{minipage}
    \begin{minipage}[t]{0.5\linewidth}
        \caption{MLP architectures}
        \label{fig:mlp_network_architecture}
    \end{minipage}%
    \begin{minipage}[t]{0.5\linewidth}
        \caption{Comparison against Neural Networks}
        \label{fig:simple_system_black_box_mse}
    \end{minipage}%
\end{figure}

\paragraph{Comparison against a Neural Network} We consider two baselines as in Fig. \ref{fig:mlp_network_architecture}. A Multi Layer Perceptron, \textbf{MLP}, takes $S_t$ as input and predicts $S_{t+1}$. MLP has 6 fully connected layers. Each layer has 30 hidden variables. The second baseline \textbf{MLP+Int} appends the integration module to MLP. It takes $S_t$ as input to predict accelerations $a_t, \alpha_t$, and then integrates for $S_{t+1}$. We sampled 1000 trajectories with different initial conditions from Mujoco. Each trajectory has 2,000 time steps. We convert the trajectories to pairs $(S_t, S_{t+1})$ for training and sample 100 more trajectories for testing. Position/Quaternion MSE in Fig.\ref{fig:simple_system_black_box_mse} means accumulated trajectory difference. \textbf{Our} method outperforms the alternatives, which fail as they do not reason about the underlying physics. \\


\section{Koopman Operator Derivation from First Principles}\label{apx:koopman_derivation}
In scenarios without external forces $f_u$, we design a physics based Koopman operator from first principles.

From equations \eqref{eq:hook}, we could infer that 
\begin{align*}
    \Vec{a} &= (f_1 + f_2) / M \\
      &= ((-K \hat{\Delta p_{1}^m} - c \hat{\Delta v_{1}^m}) + (-K \hat{\Delta p_{2}^m} - c \hat{\Delta v_{2}^m})) / M  \\
      &= (-K(\hat{\Delta p_{1}^m} + \hat{\Delta p_{2}^m}) -c ( \hat{\Delta v_{1}^m} + \hat{\Delta v_{2}^m})) / M \\
      &= -\dfrac{K}{M} (\hat{\Delta p_{1}^m} + \hat{\Delta p_{2}^m}) - \dfrac{c}{M }  ( \hat{\Delta v_{1}^m} + \hat{\Delta v_{2}^m}))
\end{align*}
Thus $\Vec{a}$ can be got with a linear function with $\hat{\Delta p_{1}^m}, \hat{\Delta p_{2}^m}, \hat{\Delta v_{1}^m}, \hat{\Delta v_{2}^m}$.

From equations \eqref{eq:2}, we could infer that 
\begin{align*}
    \Vec{\alpha} &= RI^{-1}R^{T} (r \times (f_1 - f_2)) \\
       &=  RI^{-1}R^{T} \begin{bmatrix}
            0 & -r_z & r_y \\
            r_z & 0 & -r_x \\
            -r_y & r_x & 0
          \end{bmatrix} 
          \begin{bmatrix}
            (f_1 - f_2)_x \\
            (f_1 - f_2)_y \\
            (f_1 - f_2)_z
          \end{bmatrix} \\
       &=  RI^{-1}R^{T} \begin{bmatrix}
                -r_z (f_1 - f_2)_y  + r_y (f_1 - f_2)_z \\
                r_z  (f_1 - f_2)_x  - r_x  (f_1 - f_2)_z \\
                -r_y (f_1 - f_2)_x  + r_x (f_1 - f_2)_y
          \end{bmatrix} \\
       &=  RI^{-1}R^{T} \begin{bmatrix}
                -r_z ( -K(\hat{\Delta p_{1}^m} - \hat{\Delta p_{2}^m}) - c( \hat{\Delta v_{1}^m} - \hat{\Delta v_{2}^m}))_y  + r_y ( -K(\hat{\Delta p_{1}^m} - \hat{\Delta p_{2}^m})_y - c( \hat{\Delta v_{1}^m} - \hat{\Delta v_{2}^m}))_z \\
                r_z  ( -K(\hat{\Delta p_{1}^m} - \hat{\Delta p_{2}^m}) - c( \hat{\Delta v_{1}^m} - \hat{\Delta v_{2}^m}))_x  - r_x  ( -K(\hat{\Delta p_{1}^m} - \hat{\Delta p_{2}^m})_x - c( \hat{\Delta v_{1}^m} - \hat{\Delta v_{2}^m}))_z \\
                -r_y ( -K(\hat{\Delta p_{1}^m} - \hat{\Delta p_{2}^m}) - c( \hat{\Delta v_{1}^m} - \hat{\Delta v_{2}^m}))_x  + r_x ( -K(\hat{\Delta p_{1}^m} - \hat{\Delta p_{2}^m})_x - c( \hat{\Delta v_{1}^m} - \hat{\Delta v_{2}^m}))_y
          \end{bmatrix} \\
       &= -K (RI^{-1}R^{T}) \begin{bmatrix}
                            -r_z (\hat{\Delta p_{1}^m} - \hat{\Delta p_{2}^m})_y + r_y ( \hat{\Delta p_{1}^m} - \hat{\Delta p_{2}^m})_z \\
                            r_z (\hat{\Delta p_{1}^m} - \hat{\Delta p_{2}^m})_x - r_x (\hat{\Delta p_{1}^m} - \hat{\Delta p_{2}^m})_z \\
                            -r_y (\hat{\Delta p_{1}^m} - \hat{\Delta p_{2}^m})_x + r_x (\hat{\Delta p_{1}^m} - \hat{\Delta p_{2}^m})_y
                        \end{bmatrix} \\
        & \quad  -c (RI^{-1}R^{T}) \begin{bmatrix}
                            -r_z (\hat{\Delta v_{1}^m} - \hat{\Delta v_{2}^m})_y + r_y ( \hat{\Delta v_{1}^m} - \hat{\Delta v_{2}^m})_z \\
                            r_z (\hat{\Delta v_{1}^m} - \hat{\Delta v_{2}^m})_x - r_x (\hat{\Delta v_{1}^m} - \hat{\Delta v_{2}^m})_z \\
                            -r_y (\hat{\Delta v_{1}^m} - \hat{\Delta v_{2}^m})_x + r_x (\hat{\Delta v_{1}^m} - \hat{\Delta v_{2}^m})_y                            
                       \end{bmatrix} \\
       &=  -K (RI^{-1}R^{T}) \begin{bmatrix}
                            -r_z (\hat{\Delta p_{1}^m})_y + r_z (\hat{\Delta p_{2}^m})_y + r_y ( \hat{\Delta p_{1}^m})_z - r_y (\hat{\Delta p_{2}^m})_z \\
                            r_z (\hat{\Delta p_{1}^m})_x - r_z(\hat{\Delta p_{2}^m})_x - r_x (\hat{\Delta p_{1}^m})_z + r_x (\hat{\Delta p_{2}^m})_z \\
                            -r_y (\hat{\Delta p_{1}^m})_x + r_y (\hat{\Delta p_{2}^m})_x + r_x (\hat{\Delta p_{1}^m})_y - r_x(\hat{\Delta p_{2}^m})_y
                        \end{bmatrix} \\
        & \quad -c (RI^{-1}R^{T}) \begin{bmatrix}
                            -r_z (\hat{\Delta v_{1}^m})_y + r_z (\hat{\Delta v_{2}^m})_y + r_y ( \hat{\Delta v_{1}^m})_z - r_y \hat{\Delta v_{2}^m})_z \\
                            r_z (\hat{\Delta v_{1}^m})_x - r_z (\hat{\Delta v_{2}^m})_x - r_x (\hat{\Delta v_{1}^m})_z + r_x (\hat{\Delta v_{2}^m})_z \\
                            -r_y (\hat{\Delta v_{1}^m})_x + r_y (\hat{\Delta v_{2}^m})_x + r_x (\hat{\Delta v_{1}^m})_y - r_x (\hat{\Delta v_{2}^m})_y 
                       \end{bmatrix}
\end{align*}

\begin{figure}[h]
    \centering
    \includegraphics[width=0.2\textwidth]{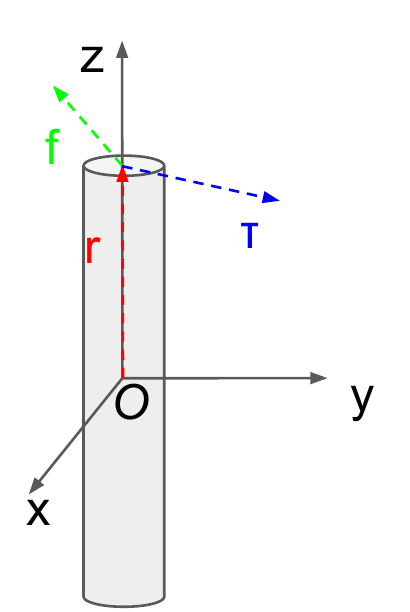}
    \caption{Torque in Rod Local Frame}
    \label{fig:local_rod_torque}
\end{figure}

Consider a uniform solid cylinder of mass $M$, radius $R$, height $2r$ as shown in Fig. \ref{fig:local_rod_torque}.
The moment of inertia of rod in local frame is 
\begin{align*}
    I = M
        \begin{bmatrix}
            \dfrac{1}{12} r^2 + \dfrac{1}{4} R^2 & 0 & 0 \\
            0 & \dfrac{1}{12} r^2 + \dfrac{1}{4} R^2 & 0 \\
            0 & 0 & \dfrac{1}{2} R^2
        \end{bmatrix}
     =  \begin{bmatrix}
            I_{11} & 0 & 0 \\
            0 & I_{11} & 0 \\
            0 & 0 & I_{33}
        \end{bmatrix}
\end{align*}
\textcolor{black}{
Since $r$ is parallel to z axis in local frame, the local frame torque $\Vec{\tau}= \Vec{r} \times \Vec{f}$ is perpendicular to z axis, i.e. $\Vec{\tau}[2]=0$.
\begin{align*}
    \Vec{\tau}_{world} &= I_{world} \Vec{\alpha} \\
    \Vec{\tau}_{world} &= R I R^T \Vec{\alpha} \\
    R^{-1} \Vec{\tau}_{world} &=  I  (R^{T}\Vec{\alpha}) \\
    \Vec{\tau} &=  I  (R^{T}\Vec{\alpha}) \\
    I^{-1} \Vec{\tau} &= R^{T}\Vec{\alpha}
\end{align*}
}
\textcolor{black}{
Since $\Vec{\tau}[2]=0$, we can get that $I^{-1} \Vec{\tau}[2] = 0$ as well no matter what $I_{33}$ is. Thus we can set $I_{33}= I_{11}$. Then we have
\begin{align*}
    RI^{-1}R^{T} =  R(I_{11}^{-1}diag(1,1,1))R^{T} = I_{11}^{-1}RR^{T} = I_{11}^{-1}E
\end{align*}
where $E$ is a 3x3 identity matrix.
}
Thus
\begin{align*}
    \Vec{\alpha} &=  
                -K I_{11}^{-1} \begin{bmatrix}
                            -r_z (\hat{\Delta p_{1}^m})_y + r_z (\hat{\Delta p_{2}^m})_y + r_y ( \hat{\Delta p_{1}^m})_z - r_y (\hat{\Delta p_{2}^m})_z \\
                            r_z (\hat{\Delta p_{1}^m})_x - r_z(\hat{\Delta p_{2}^m})_x - r_x (\hat{\Delta p_{1}^m})_z + r_x (\hat{\Delta p_{2}^m})_z \\
                            -r_y (\hat{\Delta p_{1}^m})_x + r_y (\hat{\Delta p_{2}^m})_x + r_x (\hat{\Delta p_{1}^m})_y - r_x(\hat{\Delta p_{2}^m})_y
                        \end{bmatrix} \\
        & \quad -c I_{11}^{-1} \begin{bmatrix}
                            -r_z (\hat{\Delta v_{1}^m})_y + r_z (\hat{\Delta v_{2}^m})_y + r_y ( \hat{\Delta v_{1}^m})_z - r_y \hat{\Delta v_{2}^m})_z \\
                            r_z (\hat{\Delta v_{1}^m})_x - r_z (\hat{\Delta v_{2}^m})_x - r_x (\hat{\Delta v_{1}^m})_z + r_x (\hat{\Delta v_{2}^m})_z \\
                            -r_y (\hat{\Delta v_{1}^m})_x + r_y (\hat{\Delta v_{2}^m})_x + r_x (\hat{\Delta v_{1}^m})_y - r_x (\hat{\Delta v_{2}^m})_y 
                       \end{bmatrix}
\end{align*}
Thus $\Vec{\alpha}$ is a linear combination of $r_i (\hat{\Delta v_{a}^m})_j$ and $r_i (\hat{\Delta p_{b}^m})_j$ where $i, j \in \{x, y, z\}$ and $a, b \in \{1, 2\}$.

\section{Data Efficiency Experiment on Koopman operator} \label{apx:koopman_efficiency}
Since all parameters are different, recent work compositional Koopman operator \cite{li2019learning} is no longer applicable, but our physics and topology-ware Koopman operator is still easy to scale up by adding more linear regressors. Here we apply six linear regressors for six rods. Koopman has larger error because of the non-triviality of Koopman basis functions design as we discussed in \ref{sec:tensegrity_same_params}. Especially Koopman would also fail for training on a very small dataset, 0.01\%, because the redundant basis functions may form a wrong linear function.

\begin{figure}[h]
    \centering
    \includegraphics[width=0.4\textwidth]{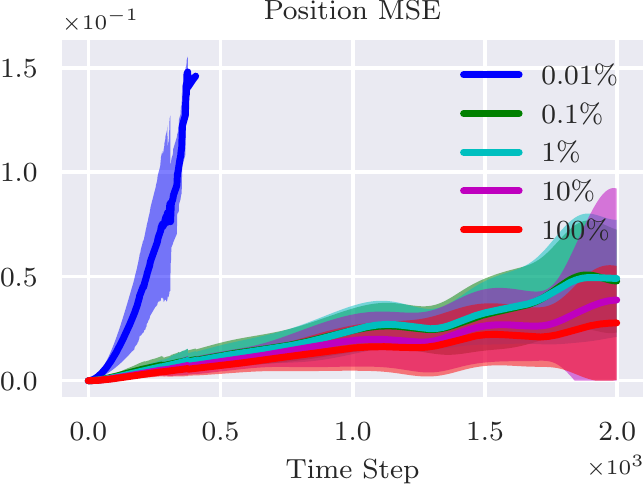}
    \includegraphics[width=0.4\textwidth]{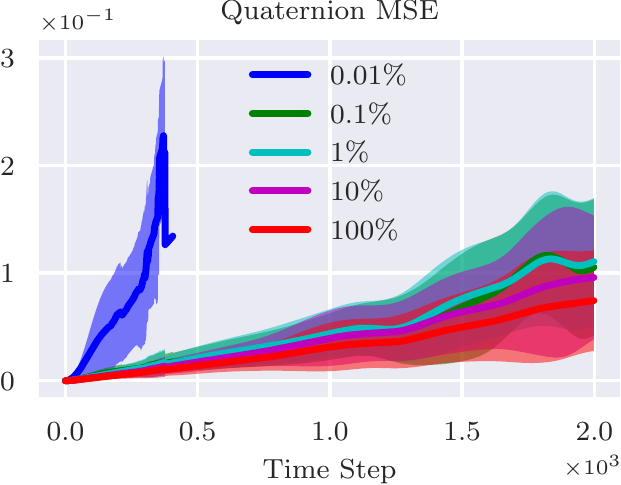}
    \caption{Position and quaternion error comparison of \textbf{Koopman operator} with different amount of training data in tensegrity model identification with \textit{different} parameters for rods and springs}
    \label{fig:superball_diff_params_system_data_efficient}
\end{figure} 

\section{Model Generalization Experiment on 4,000 Time Steps Test Set} \label{apx:generalization_4000}
\begin{figure}[h]
    \centering
    \includegraphics[width=0.4\textwidth]{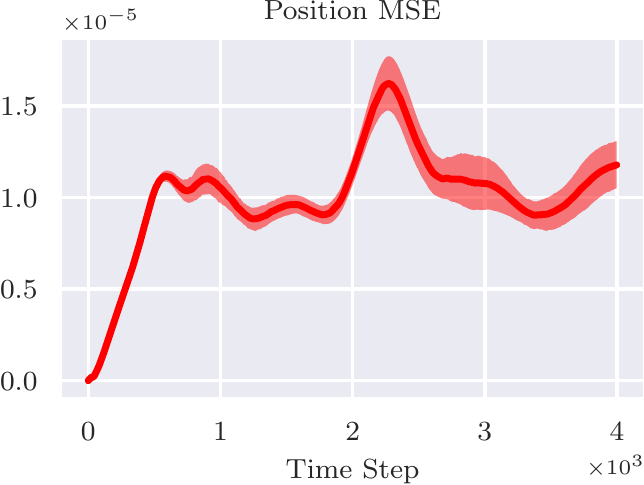}
    \includegraphics[width=0.4\textwidth]{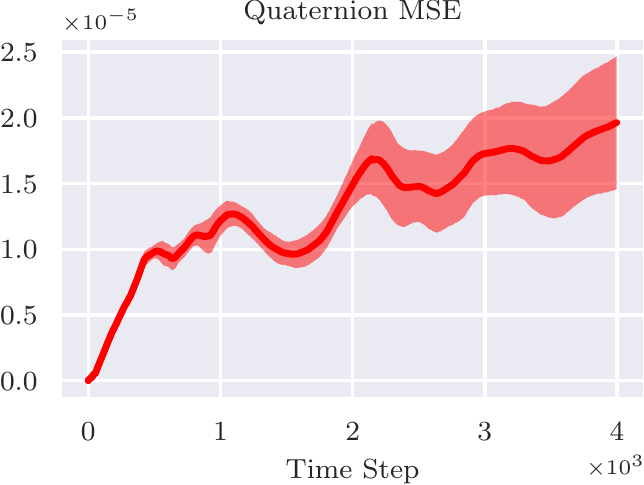}
    \caption{Position and quaternion error for 4,000 time steps test set of \textbf{Ours} tensegrity model identification with \textit{different} parameters for rods and springs, adding an arbitrary force every 100 time steps.}
    \label{fig:superball_diff_params_system_generalization}
\vspace{-0.7cm}
\end{figure}
\end{document}